!TeX program = pdflatex
\documentclass[twocolumn]{article}
\usepackage[dvipdfmx]{graphicx}
\usepackage{arxiv}
\usepackage{authblk}

\usepackage{multirow}%
\usepackage{amsmath,amssymb,amsfonts,amsthm}
\usepackage{mathrsfs}%
\usepackage[utf8]{inputenc}
\usepackage{hyperref}

\usepackage{xcolor}%
\usepackage{textcomp}%
\usepackage{manyfoot}%
\usepackage{booktabs}%
\usepackage{algorithm}%
\usepackage{algorithmicx}%
\usepackage{algpseudocode}%
\usepackage{listings}%

\usepackage{multirow}
\usepackage{subfig}
\usepackage{array}
\usepackage{tabularx}
\usepackage{threeparttable}
\usepackage{colortbl}

\theoremstyle{thmstyleone}%
\theoremstyle{thmstyletwo}%

\theoremstyle{thmstylethree}%
\newcommand{\ccol}[2]{ \multicolumn{#1}{c}{#2}}

\newcolumntype{P}[1]{>{\centering\arraybackslash}m{#1}}

\begin{document}

\newcommand{\myPaperShortTitle}{A Simple Ensemble Strategy for LLM Inference}
\newcommand{\myPaperTitle}{A Simple Ensemble Strategy for LLM Inference: \\Towards More Stable Text Classification}
\title{\myPaperTitle}
\date{}

%\author{Junichiro Niimi}

\renewcommand\Authfont{\bfseries}
\setlength{\affilsep}{0em}
% box is needed for correct spacing with authblk
\newbox{\orcid}\sbox{\orcid}{\includegraphics[scale=0.06]{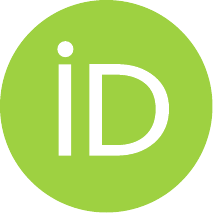}} 
\author[1,2]{%
	\href{https://orcid.org/0000-0002-4618-6272}{\usebox{\orcid}\hspace{1mm}
	Junichiro Niimi\thanks{\texttt{jniimi@meijo-u.ac.jp}}
	}}
\affil[1]{Meijo University}
\affil[2]{RIKEN AIP}

\renewcommand{\shorttitle}{\myPaperShortTitle}

\lstset{
  basicstyle={\ttfamily},
  identifierstyle={\small},
  commentstyle={\smallitshape},
  keywordstyle={\small\bfseries},
  ndkeywordstyle={\small},
  stringstyle={\small\ttfamily},
  frame={tb},
  breaklines=true,
  breakindent=0pt,
  columns=[l]{fullflexible},
  xrightmargin=0pt,
  xleftmargin=0pt,
  numberstyle={\scriptsize},
  stepnumber=1,
  numbersep=0pt,
}

%\maketitle

\twocolumn[
	\begin{@twocolumnfalse}
		\maketitle
\vspace{-3em}
\begin{abstract}
With the advance of large language models (LLMs), LLMs have been utilized for the various tasks. However, the issues of variability and reproducibility of results from each trial of LLMs have been largely overlooked in existing literature while actual human annotation uses majority voting to resolve disagreements among annotators. Therefore, this study introduces the straightforward ensemble strategy to a sentiment analysis using LLMs. As the results, we demonstrate that the ensemble of multiple inference using medium-sized LLMs produces more robust and accurate results than using a large model with a single attempt with reducing RMSE by 18.6\%. 
\end{abstract}
\vspace{0.5em}
\keywords{large language models \and natural language processing \and sentiment analysis \and marketing \and electronic word-of-mouth}
\vspace{2em}
	\end{@twocolumnfalse}
]

\renewcommand\thefootnote{*}
\setcounter{footnote}{0}
\section{Introduction}\footnote{jniimi@meijo-u.ac.jp \\This manuscript has been accepted for the 30th International Conference on Natural Language \& Information Systems (NLDB 2025) and will appear in Springer Lecture Notes in Computer Science (LNCS). Codes are available at GitHub: \url{https://github.com/jniimi/ensemble_inference}}
In the field of marketing, accurate comprehension of customer's loyalty and preference is crucial as a customer relationship management (CRM) \cite{loyalty,loyalty_measurement}. 
In particular, as consumers increasingly post opinions on social media and review platform, user-generated contents (UGCs) has become essential resource for market research \cite{nlp_in_marketing_review}. 
Textual data have been utilized for company's various decision-making, such as product evaluation, feature extraction, and recommendation systems \cite{ewom_effect,sentiment_marketing,extract_product_features,wom_to_purchase_intention,bert_hotel}. 

To extract, utilize, and understand consumer preferences from textual data, pre-processing through assigning labels is essential; however, these are labor-intensive tasks for humans. Manual labelling such as Amazon Mechanical Turk (MTurk) is costly, while traditional natural language processing (NLP) methods require specialized skills. In addition, data quality of crowdsourcing remains a serious concern \cite{mturk1,mturk2}. Thus, handling big data becomes more challenging and often impractical despite the large amount of accumulated data. 

With the advance of large language models (LLMs), several studies have proposed automated annotation models using LLMs \cite{annotation_gpt3,gptVsMTurk,annotation_gpt4}. Compared to human annotators, LLMs have highlighted advantages in processing speed, cost efficiency, and applicability across the wide range of tasks. However, the instability of LLM outputs has also been pointed out \cite{annotation_gpt4}, which can lead to inconsistency of predictions. We therefore facilitate this variability through an ensemble approach by aggregating multiple inferences to acquire both stability and accuracy. 

Therefore, this study proposes a sentiment analysis model that employs ensemble techniques, which not only stabilizes the classification but also reduces the effect of outlier generation in LLMs. The remainder of this paper is organized as follows. In Section 2, we review related works to contextualize this study. In Section 3, we introduce our proposed model, followed by an overview and results of the analysis in Section 4. Finally, we discuss the implications and  challenges of this study in Section 5. 

\section{Related Work}
First of all, sentiment analysis methods can be broadly classified into four groups: rule-based models (e.g., VADER \cite{vader}), machine learning \cite{ml-annot1,ml-annot2}, deep neural networks (DNNs) \cite{dnn_sent_review1,dnn_sent_review2}, and LLMs \cite{llm_sentiment_review}.  While many studies have shifted towards automated sentiment annotation using LLMs, as mentioned, the diversity in the responses \cite{annotation_gpt4} arises due to the nature of LLM inference, such as probabilistic token generation, prompt sensitivity, and inherent randomness. For business applications, reproducibility and consistency are crucial. In human annotation, multiple annotators are used to resolve discrepancies via majority voting, a concept that can be extended to LLM-based sentiment analysis.

Ensemble learning is typically achieved by combining multiple weak learners with diverse outputs to reduce the error rate and to generalize results \cite{ensemble}. One of the approaches of ensemble is voting, such as majority voting and weighted voting \cite{averaging}, which constructs robust predictors by combining multiple classifiers \cite{ensemble1}. Ensemble learning has been applied to machine learning \cite{ensemble_ml} and its utility also has been confirmed in the wide range of domains, including sentiment analysis \cite{ensemble1,ensemble3,ensemble_sentiment,ensemble2}. Majority voting is utilized particularly for multinomial classification, in which each model votes for one class, and the class with the most votes is considered as the final predictor \cite{ensemble2}. Given that LLMs generate probabilistic variations in their responses, ensemble learning can be realized by performing multiple inference iterations.

Despite the increasing use of LLMs, few studies have explored ensemble techniques to stabilize predictions. Most prior works have focused on single-run LLM annotations \cite{niimi_llm}. While a few studies \cite{llm_hetero,llm_hetero2} have applied ensemble-like techniques to LLM inference by assigning different characteristics to each LLM, they often rely on hierarchical models that vertically stacks LLMs, where multiple inferences are fed into another LLM for final decision-making. This approach, however, further increases the black-box nature of the model, which makes it difficult to interpret how individual predictions contribute to the final output. Consequently, little research has systematically analyzed the effect of considering the repeated inferences in LLM-based sentiment analysis. 

Therefore, this study addresses the gap by introducing a straightforward ensemble strategy to enhance stability and accuracy of the inference. We develop LLM sentiment analysis model which generates multiple workers inside the model and each worker votes the sentiment to generate more robust results. 

\section{Methodology}
\subsection{Pre-trained Models}
Table \ref{tab:models} summarizes the models in this study: Meta AI's Llama family \cite{llama3}. We adopt 70 billion (70B) and 8 billion (8B) models for Llama 3 and 7 billion (7B) model for Llama 2. Llama  is trained with more than 15 trillion tokens from publicly available data sources. Furthermore, Meta's instruction-tuning \cite{instruction-tuning} process of the model includes more than 10,000 manually annotated data in addition to instructional data, as well as reinforcement learning from human feedback (RLHF) \cite{rlhf}. Considering these learning processes, sufficient accuracy is expected for sentiment analysis of eWOM. Thus, no additional fine-tuning is performed in this study.

\begin{table*}[t] 
      \begin{center}
      \caption{Employed pre-trained models}
      \label{tab:models}
        \scalebox{1}{
\begin{threeparttable}
\begin{tabular}{
wl{5.4cm}wc{1.2cm}wr{1.4cm}wc{1.8cm}wc{1.6cm}
}
\toprule
\multicolumn{1}{c}{model id} & version & \ccol{1}{scale} & precision & format \\
\midrule
meta-llama/Llama-3-70B-Instruct & 3 & 70B~~~~ & 4-bit & GGUF \\
meta-llama/Llama-3-8B-Instruct & 3 & 8B~~~~ & 4-bit & GGUF \\
meta-llama/Llama-2-7B-Chat & 2 & 7B~~~~ & 4-bit & GGUF\\
\bottomrule
\end{tabular}
\begin{tablenotes}[para,flushleft,online,normal] %(default:normal)
The model id represents the corresponding model identifier on Hugging Face (https://huggingface.co/models).
\end{tablenotes}
\end{threeparttable}
        }%Scalebox
        \end{center}
\end{table*} 

We limit our investigation to the Llama family. Even comparing within the Llama family, we can empirically investigate the effect of the model configuration, such as the model scale (i.e., number of parameters in the model; 7--70B) and version (version 2 and 3).To reduce the computational burden, this study quantizes each model in 4-bit with GGUF format \cite{llama-cpp} which is widely adopted in LLM community. 

\subsection{Base Models}
In this study, an ensemble approach is constructed by unifying the inferences generated by different single LLMs (i.e., base models). As mentioned, since LLMs have been trained with the large-scale computing resources to ensure broad task generalizability and rapid applicability, we do not perform fine-tuning. Instead, we construct a one-shot model to ensure robust task execution. As shown in Fig. \ref{fig:prompt}, we included a single annotated example in the prompt and enabled the model to generate an accurate response. The samples used for the instructions were randomly extracted from the dataset and excluded from the test set. 
\begin{figure}[htb]
\begin{center}
\begin{lstlisting}
### Instruction
You are a helpful assistant evaluating the review texts about the restaurant. Please evaluate the review text and assign an integer score ranging from 1 for the most negative comment to 5 for the most positive comment.

### Review 1
User review: {example_review}
Output: {example_label}

### Review 2
User review: {user_review}
Output: _
\end{lstlisting}
\caption{Basic Prompt}\label{fig:prompt}
\end{center}
\end{figure}

For the text-generation task, each base model predicts a single token that follows the prompt, corresponding to the underbar at the end of the prompt. To introduce probabilistic diversity in the responses, the model's reproducibility parameter is utilized. In most machine learning methods, specifying the initial value of the random number generator (i.e., seed value), ensures reproducibility in training and inference. In other words, by repeating inferences with different seed values, the model can generate the multiple outputs as if they were produced by multiple fictitious workers. 

In this study, five virtual workers are established using different seed values. Let $LLM^w$ denote the base model using seed value $w \in \{1,2,\cdots,5\}$, and let $prompt_i$ represent the input text corresponding to sample $i \in \{1,2,\cdots,n\}$. The obtained sentiment value, denoted as $s^w_i$, is generated for sample $i$ by worker $w$. Thus, the individual output is obtained as follows:
\begin{align}
   s^w_i = LLM^w(prompt_i)
\end{align}

\subsection{Ensemble Strategy}
To implement the ensemble learning by repeating the inferences, we develop the ensemble extension of 8B model and compare the performance with non-ensemble 70B model. 
Since the target variable is ordinal scale (1--5 stars), we adopt the median for the ensemble, which is robust to outlier predictions. Majority voting, on the other hand, is commonly employed in ensemble learning; however, it is vulnerable to the case when the number of votes is tied (e.g., two votes for 1, two for 2, and one for 3). Given our focus on the diversity of LLM outputs, the median is more appropriate for the aggregation.

For sample $i \in \{1, 2, \cdots, n\}$, the unified output, denoted as $s^*_i$, is obtained as follows:
\begin{align}
   s^*_i = median(s^{w}_i | s^w_i\in\{1,2,\cdots,5\})
\end{align}
The aggregated sentiment score is computed using only valid predictions. In LLM-based sentiment analysis, models occasionally generate out-of-scope tokens, such as negative values, non-integer values, or non-digit characters. By taking the median, we mitigate the impact of such erroneous outputs and improve the robustness of the prediction.

\section{Experiments}
\subsection{Dataset Overview}
We use Yelp Open Dataset \cite{yelp}, an open dataset publicly available for academic research. Yelp is an online platform on which users post evaluations and reviews about various facilities, including restaurants, stores, and public institutions. Since our proposed method ensembles one-shot LLMs, performance is highly sensitive to pre-training data leakage. The Yelp dataset can only be obtained through an application process, so it is unlikely that an entire corpus was included for pre-training process of Llama family. 

A key challenge in analyzing Yelp Dataset is the diversity of restaurants which vary in price range, cuisine, and service. While many studies \cite{yelp0,yelp1,yelp2,yelp_nldb2024} have analyzed user reviews using this dataset due to its reproducibility, few have employed LLMs for such analyses.

The dataset includes the ratings $rating_{jk} \in \{1,2,\cdots, 5\}$ and review texts $review_{jk}$, which are posted by user $j$ about restaurant $k$, where $j \in \{1, 2, \dots, J\}$, $k \in \{1,2, \dots, K\}$. A single review was randomly selected for each user. If a user posted a review for the same restaurant multiple times, only the most recent review was considered for sampling. Thus, the sample size $n$ equals the number of users $J$. Since all users and venues are unique, sample size $n = J \times K$.

In addition, each establishment is associated with category tags; allowing us to extract the target instances based on the tags. In this study, only restaurants holding a physical store in a fixed address were included in the analysis, and therefore we extracted only those tagged with Restaurant and excluded those of Fast Food, Food Trucks, Nightlife, and Bar. We selected 1,000 instances for the test set, and Table \ref{tab:tokens} lists the summary statistics.

\begin{table}[htb] 
      \begin{center}
      \caption{Summary Statistics of the Dataset}
      \label{tab:tokens}
        \scalebox{1}{
\begin{threeparttable}
\begin{tabular}{
  wl{1.6cm}
  wr{1.2cm}wr{1.2cm} 
  wr{0.8cm} wr{0.8cm}
  }
\toprule
$n=1000$ & \ccol{1}{Mean} & \ccol{1}{Std} & \ccol{1}{Min} & \ccol{1}{Max} \\
\midrule
Characters & 392.062~~~ & 302.190~~~ & 61~~~~ & 2425~~~ \\
Tokens & 88.164~~~ & 67.973~~~ & 13~~~~ & 570~~~ \\
Stars &  3.933~~~ & 1.371~~~ & 1~~~~ & 5~~~ \\
\bottomrule
\end{tabular}
\begin{tablenotes}[para,flushleft,online,normal] %(default:normal)
The numbers of tokens are counted using the Tiktoken tokenizer \cite{tiktoken} which is also used in Llama 3.
\end{tablenotes}
\end{threeparttable}
        }%Scalebox
        \end{center}
\end{table} 

\subsection{Evaluation}
We evaluate model performance using two metrics: concordance rate (Acc.) and root mean square error (RMSE). Since the target variable is ordinal, we assess not only prediction concordance but also the magnitude of errors using RMSE. 

For the model comparison, we employ six reference models for the baseline. Among these, we implement three DNN-based models: BERT, Bi-LSTM, and CNN, based on previous studies \cite{bert,lstm_att,fasttext_cnn}. The BERT model employs bert-large-uncased for both embedding and classification, while the Bi-LSTM and CNN models incorporate word embeddings (word2vec\cite{word2vec}, FastText\cite{fasttext2}) and respective architectures. For rule-based methods, we employ VADER, which computes polarity scores directly without training. For machine learning, we adopt a TF-IDF and Linear SVM combination based on previous studies, setting the TF-IDF dimensionality to 4,000. Additionally, we establish a chance-level benchmark by estimating sentiment label distributions from the training data and generating test predictions accordingly.

\subsection{Results and Discussion}
\begin{table*}[t]
      \begin{center}
      \caption{Comparison of the accuracy and the processing time}
      \label{tab:result1}
\begin{threeparttable}
\begin{tabular}{
   wr{0.6cm}wl{3.6cm}wc{2.2cm}
   wc{1.4cm}wr{1.4cm}
   wr{1.6cm}
   }
\toprule
 \ccol{2}{Model Name} & \ccol{1}{Ensemble} & \ccol{1}{RMSE} & \ccol{1}{Acc.} & \ccol{1}{Time ($s$)} \\
\midrule
\multicolumn{2}{l}{\bf LLMs}\\
~~
1. & Llama-3-8B-Instruct & adopted & \cellcolor[gray]{0.875}\bf0.424 & \bf0.778 & 27.215 \\
2. & Llama-3-70B-Instruct & - &  \bf0.521 & \cellcolor[gray]{0.875}\bf0.779 & 64.879 \\
3. & Llama-3-8B-Instruct & - & \bf0.562 & \bf0.749 & \cellcolor[gray]{0.875}5.443\\
4. & Llama-2-7B-chat & - & \bf0.860 & \bf0.721 & 5.986 \\
\midrule[0.25pt]
\multicolumn{2}{l}{\bf Reference models}\\
5. & \multicolumn{1}{l}{BERT (Large) \cite{bert}} & - & 0.941 & 0.639 & \ccol{1}{-} \\
6. & \multicolumn{1}{l}{Bi-LSTM \cite{lstm_att}} & - & 0.930 & 0.636 & \ccol{1}{-} \\
7. & \multicolumn{1}{l}{CNN \cite{fasttext_cnn}}  & - & 1.098 & 0.596 & \ccol{1}{-} \\
8. & \multicolumn{1}{l}{VADER \cite{vader}} & - & 1.111 & \ccol{1}{~~~~~-} & \ccol{1}{-} \\
9. & \multicolumn{1}{l}{Linear SVM \cite{tfidf-svm}} & - &  1.134 & 0.627 & \ccol{1}{-} \\
10.& \multicolumn{1}{l}{Random} & - & 1.850 & 0.358 & \ccol{1}{-} \\
\midrule
\multicolumn{2}{l}{\bf Lift}\\
 & \multicolumn{2}{l}{from the best single LLM (70B)} & 18.6\% &-0.1\% & 58.1\%\\
 & \multicolumn{2}{l}{from the LLM baseline (8B)}& 32.7\% & 3.9\% &-500.0\%\\
 & \multicolumn{2}{l}{from the best reference model} & 54.4\% & 17.9\%& \ccol{1}{-} \\
\bottomrule
\end{tabular}
\begin{tablenotes}[para,flushleft,online,normal] %(default:normal)
{\it Note.} Bold metric indicates that the model performs better than all references while cell shading represents the highest performance. Time represents the average duration to process one review.
\end{tablenotes}
\end{threeparttable}
        \end{center}
\end{table*}

The results are shown in Table \ref{tab:result1}. First, non-ensemble LLMs (Model 2--4) outperformed all reference models. Notably, 70B model (Model 2) demonstrated the high performance compared to fewer-parameter models. Therefore, among the non-ensemble LLMs, the number of parameters significantly contributes on performance, which is consistent with the previous studies \cite{llm-params1}. However, 70B model (Model 2) requires approximately 11 times the processing time of the 8B model (Model 3) for only a 3\% increase in accuracy, which means that, even when analyzing only 1,000 samples, a total processing time differs by more than 16 hours. Moreover, all Llama 3 models (Models 2--3) outperformed the Llama 2 model (Model 4). Notably, Llama 2 required more inference time than 80B model (Model 3), despite having fewer parameters. These results indicate that the model architecture and pre-training process have a significant impact on both the accuracy and processing speed.

In such a situation, the proposed ensemble model even outperformed the 70B model across RMSE and almost equivalent in accuracy. This indicates that, while the ensemble model achieved similar accuracy to the 70B model, it significantly reduced the margin of error in misclassification with less than half the processing time. Thus, a medium-sized model using ensemble approach with iterative inferences is significantly more efficient in terms of both training time and model performance than the larger model with a single inference.

Across the non-LLM baselines (Models 5--10), although the DNN-based models (Models 5--7) achieved high accuracy as expected, none of them surpassed LLM-based models. Additionally, VADER (Model 8) outperformed the machine-learning model (Model 9). Nevertheless, all models performed above the chance level (Model 10).

\section{Conclusion}
In this study, we demonstrated the effectiveness of a simple ensemble approach for sentiment analysis using LLMs. With the proposed model, which imitates human annotation, the results demonstrated a significant improvement over non-ensemble LLMs. By introducing the ensemble approach, prediction errors are compensated by each model, which significantly improves the accuracy (Table \ref{tab:voting}). 
Thus, the proposed approach enables a medium-sized model to achieve performance comparable to a 70B model while requiring less than half the processing time. Moreover, even non-ensemble LLMs, compared with the well-known existing methods, can perform higher accuracy without any additional training. Thus, LLMs are highly useful for practical applications due to their large-scale pre-training.

\newcommand{\colwid}{1.2cm}
\begin{table*}[t]
  \begin{center}
  \caption{The Effects of Ensemble Inference}\label{tab:voting}
    \begin{tabular}{
    wc{0.4 cm} wl{2.0 cm} 
    wc{\colwid}wc{\colwid}wc{\colwid}wc{\colwid}wc{\colwid}
    wc{\colwid}wc{\colwid}wc{\colwid}wc{\colwid}
    }
       \toprule 
       \ccol{2}{\multirow{2}{*}{Sample}}&\ccol{5}{Workers $w \in \{1,2,\cdots,5\}$}&\multirow{2}{*}{$s_i^*$} \\%&\multirow{2}{*}{Label}\\
       \cmidrule{3-7} 
        && $s_i^1$ & $s_i^2$ & $s_i^3$ & $s_i^4$ & $s_i^5$  \\
        \midrule
        1. & consistent & 4 & 4 & 4 & 4 & 4  & 4 \\%& 4 \\
        2. & similar      & 2 & 2 & 2 & 3 & 3 & 2 \\%& 2\\
        3. & outlier  & 5 & 5 & 5 & 5 & 1 & 5 \\%& 5\\
       \bottomrule
    \end{tabular}
  \end{center}
\end{table*}

This study has both academic and practical implications. First, from an academic perspective, our study suggests that simple ensemble methods can enhance the model performance using the randomness of LLM outputs. Unlike the other ensemble methods which rely on multiple independently trained models (e.g., bagging, boosting), our approach is based on the stochastic nature of LLM inferences. This indicates that this strategy is generally applicable to other pre-trained LLMs with different architecture or different number of parameters. Also, from a practical perspective, our findings provide two key implications. First, the ensemble of multiple inferences with a medium-sized model are much faster and more accurate than a single inference with a larger model. This efficiency can enhance the understanding of consumer preference and support quick decision-making for business. Second, since ensemble inference does not require additional training costs, practitioners can readily improve the tasks with this approach in their business environments.

Finally, this study remains several future challenges. First, all the base models employed the same example in the prompt; however, the performance of LLMs is highly sensitive to the prompt. Future research should explore the effects of different prompt configurations across the base models. Second, while this study focuses on ensemble of multiple models, it is necessary to examine the effect of combining different-scale models. This study treated all the outputs equally by using the median; however, to effectively integrate models with varying parameter counts, alternative ensemble techniques such as weighted averaging may be required. Third, although this study applied the ensemble of LLM inferences to marketing analysis on restaurant reviews, the proposed ensemble strategy itself is applicable to general text classification tasks. Future work should evaluate its effectiveness on broader benchmark across different domains to further validate the generalizability.

\bibliographystyle{unsrt}
\bibliography{NLDB2025}

\end{document}